\documentclass[11pt]{article}
\usepackage[preprint]{acl}

\usepackage{times}
\usepackage{latexsym}
\usepackage[T1]{fontenc}
\usepackage[utf8]{inputenc}
\usepackage{microtype}
\usepackage{inconsolata}
\usepackage{graphicx}
\usepackage{amsmath,amssymb}
\usepackage{booktabs}
\usepackage{tabularx}
\usepackage{colortbl}
\definecolor{oursblue}{RGB}{228,237,250}
\newcommand{\oursrow}{\rowcolor{oursblue}}
\usepackage{multirow}
\usepackage{xcolor}
\usepackage{enumitem}
\usepackage{xspace}
\usepackage{algorithm}
\usepackage[noend]{algpseudocode}
\usepackage[most]{tcolorbox}
\definecolor{promptframe}{RGB}{200,204,210}
\definecolor{promptback}{RGB}{250,250,248}
\definecolor{prompttitle}{RGB}{236,238,241}
\newtcolorbox{promptbox}[1]{enhanced, colback=promptback,
  colframe=promptframe, colbacktitle=prompttitle, coltitle=black!80,
  boxrule=0.5pt, titlerule=0.5pt, arc=3pt, left=7pt, right=7pt,
  top=5pt, bottom=5pt, toptitle=2pt, bottomtitle=2pt, title={#1},
  fonttitle=\small\bfseries, before skip=8pt, after skip=8pt}

\newcommand{\ours}{MRTPO\xspace}
\newcommand{\JS}{\mathrm{JS}}

\title{Each Reward to Its Relevant Tokens:\\Reward-Specific Token-Level Credit Assignment for Multi-Reward RL}

\author{
 \textbf{Skylar Zhai\textsuperscript{$\alpha$}},
 \textbf{Haonan Wang\textsuperscript{$\beta$}},
 \textbf{Zhining Liu\textsuperscript{$\gamma$}},
 \textbf{Linxin Song\textsuperscript{$\delta$}},
\\
 \textbf{Xuwei Ding\textsuperscript{$\varepsilon$}},
 \textbf{Jingcheng Liang\textsuperscript{$\alpha$}},
 \textbf{Jian Kang\textsuperscript{$\zeta$}}
\\
 \textsuperscript{$\alpha$}University of Minnesota \quad
 \textsuperscript{$\beta$}National University of Singapore \quad
 \textsuperscript{$\gamma$}University of Illinois Urbana-Champaign
\\
 \textsuperscript{$\delta$}University of Southern California \quad
 \textsuperscript{$\varepsilon$}University of Pennsylvania \quad
 \textsuperscript{$\zeta$}MBZUAI
\\
 \small{
   \textbf{Code:} \url{https://github.com/zhaihaotian/MRTPO} \quad
   \textbf{Project page:} \url{https://zhaihaotian.github.io/MRTPO/}
 }
}

\begin{document}
\maketitle

\begin{abstract}
Language models are increasingly trained against several verifiable rewards at
once, such as the correctness and the format of a tool call, or the individual
constraints of a complex instruction. Reward-decoupled normalization (GDPO)
normalizes each reward separately so that different reward combinations keep
distinct advantages, but it still spreads every reward's advantage evenly over
the whole response, although each verifier checks only a few positions. We
propose \ours, which assigns the advantage of each reward to the tokens
relevant to it, using the information its verifier checks, without auxiliary
models or token-level annotations. For each reward, the
policy replays its own response with and without that reward's verifier
information, and the per-token Jensen--Shannon distance between the two
next-token distributions measures each token's relevance and reassigns the
advantage. On text-to-SQL, multi-constraint instruction following and tool
calling, \ours converges faster and achieves better performance than the
compared methods.
\end{abstract}

\section{Introduction}

\begin{figure}[t]
  \centering
  \includegraphics[width=\columnwidth]{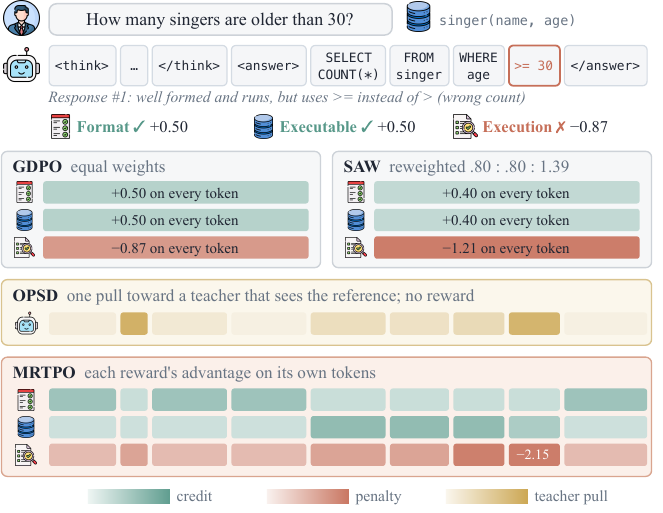}
  \caption{\textbf{Advantages assigned to the tokens of one SQL response} that
is well formed and executable but uses \texttt{>=} for \texttt{>}. \ours places
format credit on the tags, executability credit on the query, and the
execution penalty mostly on \texttt{>= 30}.}
  \label{fig:teaser}
\end{figure}

Reinforcement learning with verifiable rewards (RLVR) has become a standard component of large language model (LLM) post-training, with Group Relative Policy Optimization (GRPO) and its variants among the most widely adopted algorithms~\citep{shao2024deepseekmath,guo2025deepseekr1}. 
Many practical objectives, however, involve multiple criteria. 
A tool-using agent must emit a well-formed call and also select the correct function and arguments~\citep{qian2025toolrl}; 
a text-to-SQL model must produce a query that both executes and returns the intended result~\citep{ma2025sqlr1}; 
an instruction-following model must satisfy every constraint stated in the prompt~\citep{zhou2023ifeval}; 
and an aligned assistant must be both helpful and harmless~\citep{dai2024saferlhf}.
Each criterion is typically evaluated by a separate verifier that produces its own reward, which gives rise to multi-reward reinforcement learning~\citep{liu2026gdpo}.

Applying GRPO to the sum of these rewards collapses distinct reward combinations into identical advantage values \citep{liu2026gdpo}. 
To address this reward collapse, GDPO normalizes each reward within its rollout group before aggregation \citep{liu2026gdpo}, and a line of concurrent work further adapts the weight of each reward during aggregation \citep{jiang2026dvao,he2026saw,wang2026samrpo,liu2026gd2po,montero2026rvpo}.
These methods determine how much each reward contributes to the response-level advantage, yet the resulting advantage is still assigned uniformly to every token of the response. 
This implicitly treats every token as equally responsible for every reward, whereas a verifier typically evaluates only a small portion of the response. 
For instance, if a tool call specifies all arguments correctly but omits the required \texttt{<think>} tag, the format penalty is imposed equally on the correctly generated arguments. 
This motivates the following question: \textbf{\emph{Can we identify the tokens relevant to each reward and assign the reward’s advantage mainly to them?}} 

Recent work on on-policy self-distillation suggests an answer. 
Conditioning a model on privileged information, such as a reference solution or environment feedback, alters its next-token predictions on a response it has already generated; 
self-distillation methods train the policy to match these privileged predictions, thereby replacing reward maximization with distillation toward a teacher distribution \citep{zhao2026opsd,hubotter2026sdpo}. In multi-reward RL, such information is available separately for each reward, because every verifier holds the information it relies on for evaluation, such as a reference tool call, the execution error of a generated query, or an individual constraint of the instruction. 
Since this information decides the reward, we call a token relevant to it when the information
changes its prediction, and distribute the reward's advantage by this relevance.

Building on this insight, we propose \ours (\textbf{M}ulti-\textbf{R}eward-to-\textbf{T}oken \textbf{P}olicy \textbf{O}ptimization), a token-level credit assignment method for multi-reward RL. 
For each reward, \ours 
replays the sampled response under two contexts that differ only in the information of that reward's verifier, computes the token-wise Jensen--Shannon distance between the resulting next-token distributions, and redistributes the group-normalized advantage of the reward across tokens according to this distance. 
This redistribution is what separates \ours from prior methods. GDPO~\citep{liu2026gdpo} and SAW~\citep{he2026saw}
decide how much each reward contributes to a response but give every token the
same share, whereas \ours decides which tokens receive each reward's advantage.
OPSD \citep{zhao2026opsd} also uses privileged information and trains the policy to fit a teacher that sees it. We highlight these differences in Figure~\ref{fig:teaser}.
Our main contributions are as follows:
\begin{itemize}[leftmargin=*, nosep]
  \item \textbf{A credit-assignment perspective on multi-reward RL.} We show that existing multi-reward methods determine the weight of each reward only at the response level and assign 
  the advantage of every reward 
  uniformly to all tokens, although only part of the response is relevant to each reward.
  \item \textbf{Reward-specific token-level credit assignment.} We propose \ours, which constructs context pairs from each verifier's own information and derives token weights for each reward separately through self-replay, requiring no auxiliary model, token-level annotation or additional generation.
  \item \textbf{Experiments.} Experiments on text-to-SQL, multi-constraint
instruction following and tool calling show that \ours achieves higher
performance than GDPO with faster convergence.
\end{itemize}

\section{Related Work}
\label{sec:related}
\noindent \textbf{Multi-reward reinforcement learning.}
Early work on multi-objective alignment balances several objectives by training
separate reward models or by interpolating policies trained on different
rewards \citep{dai2024saferlhf,rame2023rewardedsoups,wu2023finegrained}. Within
the GRPO framework \citep{shao2024deepseekmath}, the question becomes how to
turn a reward vector into group-relative advantages. GDPO normalizes each
reward within its group before summation so that different reward combinations
remain distinguishable \citep{liu2026gdpo}. Concurrent work builds on GDPO and
adjusts the weight of each reward dynamically according to within-group variance,
training progress or conflicts between rewards
\citep{jiang2026dvao,he2026saw,wang2026samrpo,liu2026gd2po,montero2026rvpo,zheng2026makedara}. All of these methods operate at the response level. 
Different from these approaches, \ours keeps the reward-wise group advantages
of GDPO and further decides which tokens receive each reward's advantage, so
it can be combined with any of these reweighting methods, as we do with SAW
in Section~\ref{sec:setup}.

\noindent \textbf{Token-level credit assignment and privileged information.}
To obtain credit finer than a response-level advantage, prior work relies on
process reward models \citep{lightman2023verify}, Monte-Carlo value estimates
\citep{kazemnejad2024vineppo} or token-level discriminators trained for each
rubric item \citep{rtt2026}, which require step-level annotations, additional
sampling or an additional model. Other methods derive token weights from token
entropy \citep{wang2025highentropy,cheng2025entropy,wang2025empg} or from
replaying the response without the entire rubric
\citep{zhang2026cort}, but they reweight the aggregated advantage rather than
each reward separately.
Another line of work uses privileged information for on-policy self-distillation. Given a reference solution or environment feedback, the same model acts as the teacher, and the student fits the distribution of the teacher~\citep{agarwal2024gkd,zhao2026opsd,hubotter2026sdpo}.
\ours also builds on privileged information and self-replay, and uses the
distributional difference under each reward's information to allocate that
reward's advantage over tokens. The rewards thus remain the training
objective, and the replay only decides where each advantage goes.

\section{Preliminaries}
\label{sec:prelim}
\noindent \textbf{Multi-reward group optimization.}
For each prompt $x$ in a training batch $\mathcal{B}$, the behavior policy
$\pi_{\theta_{\mathrm{old}}}$ samples a group of $G$ responses
$y^{(1)},\dots,y^{(G)}$, where $y^{(j)}$ has $T_j$ tokens. Each response is
scored by $K$ verifiers, giving rewards $r^{(j)}_1,\dots,r^{(j)}_K$; the index
$k$ always refers to a reward. $K$ may vary across prompts; in instruction
following, for example, every constraint in the prompt is a reward. We write
group standardization as
\begin{equation}
\mathcal{G}\big(r^{(j)}_k\big)=\frac{r^{(j)}_k-\mu_k}{\sigma_k+\epsilon},
\end{equation}
where $\mu_k$ and $\sigma_k$ are the mean and sample standard deviation of
reward $k$ over the $G$ responses.

\noindent \textbf{GRPO and GDPO.}
GRPO aggregates the rewards before standardizing them,
$A^{(j)}=\mathcal{G}\big(\sum_k r^{(j)}_k\big)$ \citep{shao2024deepseekmath},
so responses with different reward combinations but equal sums receive the same advantage. 
GDPO standardizes each reward and 
takes a weighted sum \citep{liu2026gdpo}:
\begin{equation}
A^{(j)}_k=\mathcal{G}\big(r^{(j)}_k\big),\qquad
S^{(j)}=\sum_{k=1}^{K}c_k\,A^{(j)}_k,
\label{eq:gdpo}
\end{equation}
followed by a batch-level rescaling
$A^{(j)}_{\textsc{gdpo}}=S^{(j)}/\mathrm{std}_{\mathcal{B}}(S)$. GDPO uses
$c_k=1$. Most concurrent work keeps this form and sets or adapts the reward
weights $c_k$ according to within-group variance, training progress or
conflicts between rewards
\citep{jiang2026dvao,he2026saw,wang2026samrpo,liu2026gd2po,montero2026rvpo}.
We use $c_k=1$ and omit it below.

\noindent \textbf{Token-level objective.}
In existing methods, whatever $c_k$ is, all tokens of response $j$ share one advantage, $A^{(j)}_t = A^{(j)}$ for every $t$.
It enters the clipped surrogate objective
\begin{equation}
\begin{aligned}
\mathcal{L}(\theta)&=-\frac{1}{\sum_j T_j}\sum_{j}\sum_{t=1}^{T_j}\ell^{(j)}_t,\\
\ell^{(j)}_t&=\min\!\big(\rho^{(j)}_t A^{(j)}_t,\;\bar\rho^{(j)}_t A^{(j)}_t\big),
\end{aligned}
\label{eq:ppo}
\end{equation}
where $\rho^{(j)}_t$ is the probability ratio between the current and the
behavior policy at token $t$ and
$\bar\rho^{(j)}_t=\mathrm{clip}(\rho^{(j)}_t,1-\varepsilon_l,1+\varepsilon_h)$.
\ours changes only $A^{(j)}_t$ and leaves the rest of the objective intact.

\section{\ours: Assigning Credit of Each Reward to Relevant Tokens }
\label{sec:method}
\subsection{Motivation}
\label{sec:motivation}

\begin{figure}[t]
  \centering
  \includegraphics[width=\columnwidth]{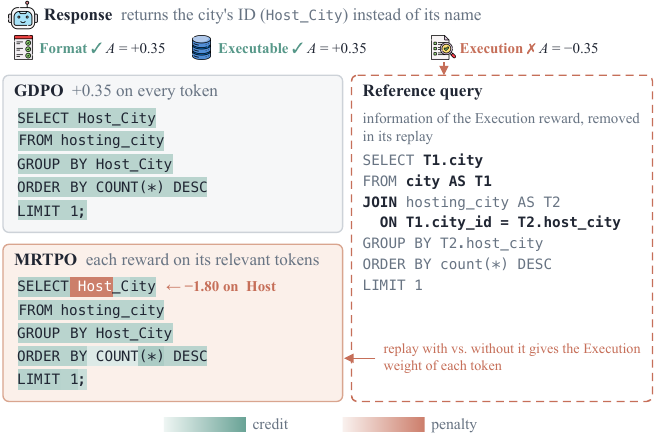}
  \caption{\textbf{Per-token advantage of a training response under GDPO and
  \ours.} Both panels use the same per-reward advantages.}
  \label{fig:sql-credit}
\end{figure}

Because every token shares one advantage (Eq.~\ref{eq:ppo}), the penalty for an
error is spread over the whole response and can be outweighed by the credit of
other rewards. In Figure~\ref{fig:sql-credit}, only the selected column is
wrong, yet GDPO credits the entire query, the wrong column included. Each
reward is decided by a few tokens, here the execution reward by the selected
column. \ours finds the tokens relevant to each reward
(Section~\ref{sec:replay}) and places the reward's advantage mainly on them
(Section~\ref{sec:allocation}).

\subsection{How Do We Find Relevant Tokens?}
\label{sec:replay}

\begin{figure*}[t]
  \centering
  \includegraphics[width=\textwidth]{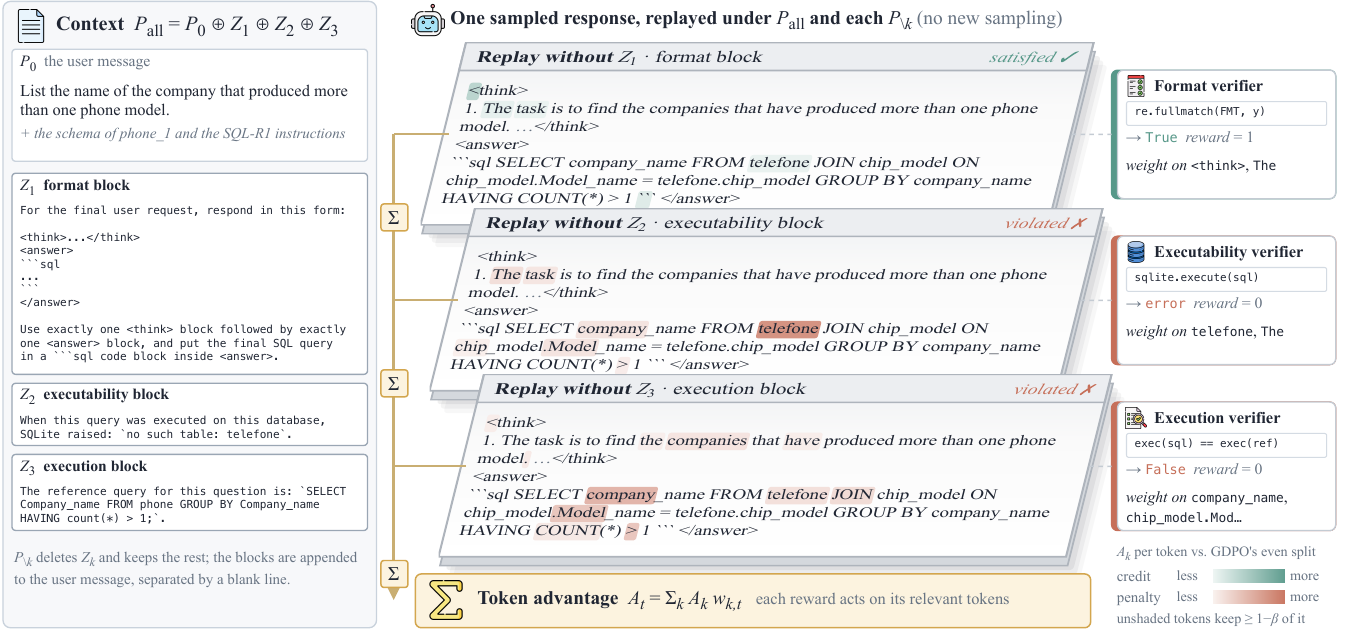}
  \caption{\textbf{Leave-one-out replay on a text-to-SQL response.} The response
  reads table \texttt{telefone}, which does not exist; the reference uses
  \texttt{phone}. Replay places the penalties of the executability and
  execution rewards mainly on \texttt{telefone} and the selected columns, and
  the format credit on the tags.}
  \label{fig:replay-sql}
\end{figure*}

By self-replay. For each reward, \ours builds two contexts that differ only
in the information of the reward's verifier, replays the response under both,
and scores each token by how much the two predictions differ. The score serves
as the token relevance to the reward.

\noindent \textbf{Verifier information.}
Each verifier scores a response against some information, such as a reference
answer or the requirement it checks. We write this information for reward $k$
as a text block $Z_k$. In Figure~\ref{fig:replay-sql} (left), $Z_1$ states the
required format, $Z_2$ reports the SQLite error of the query, and $Z_3$ gives
the reference query. The blocks of each task are listed in
Appendix~\ref{app:contexts}.

\noindent \textbf{Leave-one-out contexts.}
Let $P_0$ be the prompt that sampled $y$. Both contexts are built from $P_0$
and the blocks, so sampling stays as in GDPO. Let $\mathcal{K}_x$ be the
rewards with a block for sample $x$, $Z_{\mathcal{S}}$ the blocks in
$\mathcal{S}$ concatenated in a fixed order, and $\oplus$ appending text to the
user message. For each reward $k$,
\begin{equation}
P_{\mathrm{all}}=P_0\oplus Z_{\mathcal{K}_x},\qquad
P_{\setminus k}=P_0\oplus Z_{\mathcal{K}_x\setminus\{k\}} .
\label{eq:contexts}
\end{equation}
$P_{\mathrm{all}}$ appends the blocks of all rewards to the prompt, and
$P_{\setminus k}$ appends all of them except $Z_k$. The two contexts differ
only in $Z_k$, so a change in prediction between them comes from what $Z_k$
says. Comparing $P_0$ with $P_0\oplus Z_k$ would also capture the effect of
extra text. $P_{\mathrm{all}}$ is replayed once for all rewards.
When the information is in the prompt, as for instruction constraints, $P_{\mathrm{all}}$ is the prompt itself and $P_{\setminus k}$
removes one constraint (Figure~\ref{fig:hir-example}).

\noindent \textbf{Self-replay.}
With the parameters $\theta_{\mathrm{old}}$ that sampled $y$, we feed $y$
after each context and read the next-token distributions $p^{\mathrm{all}}_t$
under $P_{\mathrm{all}}$ and $p^{\setminus k}_t$ under $P_{\setminus k}$ at
every position $t$. Each replay is one forward pass over the sampled response.

\noindent \textbf{Relevance score.}
The more the two distributions differ at position $t$, the more the prediction
there depends on $Z_k$. We score token $t$ for reward $k$ by the Jensen--Shannon
distance 
between them,
normalized by $\ln 2$,
\begin{equation}
q_{k,t}=\sqrt{\JS\big(p^{\mathrm{all}}_t\,\big\|\,p^{\setminus k}_t\big)/\ln 2}\;\in[0,1],
\label{eq:q}
\end{equation}
which is $0$ when removing $Z_k$ leaves the prediction at $t$ unchanged
(Appendix~\ref{app:weighting}). In Figure~\ref{fig:replay-sql}, removing the
SQLite error changes the prediction mostly at \texttt{telefone}, the table
that does not exist, and removing the reference query changes it at the
selected column and the join condition.

\subsection{Redistributing Advantage by Relevance}
\label{sec:allocation}

\begin{table*}[t]
\centering
\small
\setlength{\tabcolsep}{3pt}
\renewcommand{\arraystretch}{1.1}
\begin{tabularx}{\textwidth}{@{}l*{7}{>{\centering\arraybackslash}X}@{}}
\toprule
& \multicolumn{5}{c}{Spider} & BIRD & \\
\cmidrule(lr){2-6}\cmidrule(lr){7-7}
Method & Dev & Test & Syn & Realistic & DK & Dev & Average \\
\midrule
\multicolumn{8}{@{}l}{\emph{Qwen2.5-Coder-3B-Instruct}} \\
Base & 62.38 & 64.46 & 53.00 & 54.72 & 51.87 & 36.88 & 53.88 \\
GRPO & 74.24$_{\pm1.26}$ & 75.00$_{\pm0.31}$ & 65.92$_{\pm1.50}$ & 69.49$_{\pm1.94}$ & 64.04$_{\pm1.69}$ & 43.54$_{\pm1.03}$ & 65.37$_{\pm1.18}$ \\
GDPO & 76.66$_{\pm2.21}$ & 76.40$_{\pm1.17}$ & 67.34$_{\pm1.82}$ & 70.80$_{\pm1.94}$ & 65.61$_{\pm0.93}$ & \textbf{47.13}$_{\pm1.94}$ & 67.32$_{\pm1.66}$ \\
OPSD & 68.18$_{\pm0.61}$ & 69.34$_{\pm0.45}$ & 59.15$_{\pm0.39}$ & 62.14$_{\pm1.38}$ & 56.12$_{\pm0.96}$ & 39.21$_{\pm0.70}$ & 59.02$_{\pm0.49}$ \\
\oursrow \ours & \textbf{78.24}$_{\pm0.44}$ & \textbf{76.79}$_{\pm0.43}$ & \textbf{68.67}$_{\pm0.86}$ & \textbf{73.49}$_{\pm0.61}$ & \textbf{68.17}$_{\pm1.53}$ & 46.26$_{\pm0.87}$ & \textbf{68.60}$_{\pm0.15}$ \\
\arrayrulecolor{black!30}\midrule\arrayrulecolor{black}
GDPO + SAW & 74.63$_{\pm0.06}$ & 74.77$_{\pm0.86}$ & 65.60$_{\pm1.03}$ & 70.60$_{\pm2.08}$ & 64.67$_{\pm0.58}$ & 45.23$_{\pm1.28}$ & 65.92$_{\pm0.93}$ \\
\oursrow \ours + SAW & \textbf{78.88}$_{\pm0.44}$ & \textbf{77.67}$_{\pm0.23}$ & \textbf{68.86}$_{\pm0.54}$ & \textbf{73.49}$_{\pm0.57}$ & \textbf{67.92}$_{\pm0.66}$ & \textbf{46.95}$_{\pm0.93}$ & \textbf{68.96}$_{\pm0.19}$ \\
\midrule
\multicolumn{8}{@{}l}{\emph{Qwen2.5-Coder-7B-Instruct}} \\
Base & 77.18 & 77.55 & 68.67 & 73.62 & 67.04 & 51.63 & 69.28 \\
GRPO & 78.30$_{\pm0.78}$ & 81.03$_{\pm0.49}$ & 71.44$_{\pm0.44}$ & 73.89$_{\pm0.79}$ & 66.91$_{\pm0.66}$ & 54.00$_{\pm0.63}$ & 70.93$_{\pm0.32}$ \\
GDPO & 79.21$_{\pm0.68}$ & 80.62$_{\pm0.49}$ & 72.21$_{\pm0.73}$ & 74.87$_{\pm0.82}$ & 67.29$_{\pm0.54}$ & 55.11$_{\pm1.26}$ & 71.55$_{\pm0.27}$ \\
OPSD & 79.11$_{\pm0.44}$ & 80.39$_{\pm0.52}$ & 70.76$_{\pm1.45}$ & 75.72$_{\pm0.97}$ & 69.10$_{\pm0.49}$ & 53.87$_{\pm0.36}$ & 71.49$_{\pm0.54}$ \\
\oursrow \ours & \textbf{81.33}$_{\pm0.42}$ & \textbf{81.65}$_{\pm0.45}$ & \textbf{73.53}$_{\pm0.48}$ & \textbf{75.92}$_{\pm1.78}$ & \textbf{70.04}$_{\pm1.79}$ & \textbf{56.42}$_{\pm0.46}$ & \textbf{73.15}$_{\pm0.19}$ \\
\arrayrulecolor{black!30}\midrule\arrayrulecolor{black}
GDPO + SAW & 79.92$_{\pm0.46}$ & 81.32$_{\pm0.47}$ & 72.24$_{\pm0.98}$ & 76.31$_{\pm0.12}$ & 68.23$_{\pm1.56}$ & 55.44$_{\pm1.18}$ & 72.24$_{\pm0.35}$ \\
\oursrow \ours + SAW & \textbf{81.08}$_{\pm0.31}$ & \textbf{81.39}$_{\pm0.15}$ & \textbf{73.21}$_{\pm1.12}$ & \textbf{77.23}$_{\pm0.99}$ & \textbf{70.04}$_{\pm0.19}$ & \textbf{57.53}$_{\pm0.32}$ & \textbf{73.41}$_{\pm0.35}$ \\
\bottomrule
\end{tabularx}
\caption{Text-to-SQL execution accuracy (\%); subscripts are standard deviations over seeds. \ours is best on average with both models, with and without SAW.}
\label{tab:sql}
\end{table*}

\begin{table*}[t]
\centering
\footnotesize
\setlength{\tabcolsep}{0pt}
\renewcommand{\arraystretch}{1.0}
\begin{tabular*}{\textwidth}{@{\extracolsep{\fill}}l*{8}{c}@{}}
\toprule
& \multicolumn{2}{c}{IFEval} & \multicolumn{2}{c}{IFBench} & MulDimIF & \multicolumn{2}{c}{AdvancedIF} & \\
\cmidrule(lr){2-3}\cmidrule(lr){4-5}\cmidrule(lr){6-6}\cmidrule(lr){7-8}
Method & Prompt & Inst. & Prompt & Inst. & Acc. & Overall & Rubric & Avg. \\
\midrule
\multicolumn{9}{@{}l}{\emph{Qwen3-4B-Instruct-2507}} \\
Instruct & 83.77 & 88.82 & 30.95 & 33.13 & 56.50 & 38.21 & 76.41 & 58.26 \\
GRPO & 85.44$_{\pm0.15}$ & 90.16$_{\pm0.06}$ & 35.68$_{\pm1.02}$ & 37.79$_{\pm1.01}$ & 62.44$_{\pm0.86}$ & 39.27$_{\pm0.27}$ & 77.06$_{\pm0.12}$ & 61.12$_{\pm0.34}$ \\
CoRT & 85.37$_{\pm0.50}$ & 90.06$_{\pm0.31}$ & 34.74$_{\pm1.90}$ & 36.85$_{\pm1.95}$ & 63.00$_{\pm1.35}$ & 38.88$_{\pm0.39}$ & 76.95$_{\pm0.44}$ & 60.84$_{\pm0.78}$ \\
GDPO & 85.52$_{\pm0.52}$ & 90.23$_{\pm0.33}$ & 36.39$_{\pm1.59}$ & 38.84$_{\pm1.18}$ & 62.88$_{\pm2.83}$ & 38.91$_{\pm0.43}$ & 76.85$_{\pm0.21}$ & 61.37$_{\pm0.77}$ \\
\oursrow \ours & \textbf{85.66}$_{\pm1.20}$ & \textbf{90.40}$_{\pm0.74}$ & \textbf{38.76}$_{\pm1.53}$ & \textbf{40.92}$_{\pm1.33}$ & \textbf{65.37}$_{\pm1.33}$ & \textbf{39.39}$_{\pm0.61}$ & \textbf{77.25}$_{\pm0.15}$ & \textbf{62.54}$_{\pm0.32}$ \\
\midrule
\multicolumn{9}{@{}l}{\emph{Qwen2.5-7B-Instruct}} \\
Instruct & 72.09 & 79.50 & 26.53 & 29.37 & 50.77 & 23.51 & 69.26 & 50.15 \\
GRPO & 81.72$_{\pm0.31}$ & 87.42$_{\pm0.16}$ & 35.94$_{\pm2.06}$ & 38.43$_{\pm2.05}$ & 55.19$_{\pm0.77}$ & 25.98$_{\pm0.56}$ & 71.08$_{\pm0.24}$ & 56.54$_{\pm0.62}$ \\
CoRT & 81.32$_{\pm0.56}$ & 87.11$_{\pm0.40}$ & 35.68$_{\pm1.57}$ & 38.22$_{\pm1.73}$ & 55.05$_{\pm0.30}$ & 26.07$_{\pm0.76}$ & 71.11$_{\pm0.37}$ & 56.36$_{\pm0.55}$ \\
GDPO & 81.91$_{\pm1.30}$ & \textbf{87.58}$_{\pm0.94}$ & 35.70$_{\pm2.63}$ & 38.57$_{\pm3.25}$ & \textbf{55.33}$_{\pm1.58}$ & 26.13$_{\pm0.90}$ & \textbf{71.46}$_{\pm0.25}$ & 56.67$_{\pm0.77}$ \\
\oursrow \ours & \textbf{81.98}$_{\pm0.78}$ & 87.57$_{\pm0.60}$ & \textbf{38.87}$_{\pm1.96}$ & \textbf{42.00}$_{\pm2.60}$ & 53.50$_{\pm1.72}$ & \textbf{26.18}$_{\pm0.52}$ & 71.44$_{\pm0.33}$ & \textbf{57.36}$_{\pm0.63}$ \\
\bottomrule
\end{tabular*}
\caption{Instruction-following accuracy (\%); subscripts are standard deviations over seeds. \ours has the best average on both models, and its largest gains are on IFBench, whose constraints are not seen in training.}
\label{tab:hir}
\end{table*}
With the relevance scores, each reward's advantage is redistributed over the
tokens according to their relevance relative to the response average
$\bar q_k=\frac1T\sum_u q_{k,u}$:
\begin{equation}
w_{k,t}=(1-\beta)+\beta\,\frac{q_{k,t}}{\bar q_k},
\label{eq:weight}
\end{equation}
where $\beta\in[0,1)$ is the share of the advantage that follows relevance. A
token of average relevance keeps weight $1$, a more relevant token receives
more, and every token keeps at least $1-\beta$. The rewards are then summed
per token,
\begin{equation}
S^{(j)}_t=\sum_{k=1}^{K}A^{(j)}_k\,w^{(j)}_{k,t},\qquad
A^{(j)}_{t}=\frac{S^{(j)}_t}{\mathrm{std}_{\mathcal{B}}(S)} ,
\label{eq:token-adv}
\end{equation}
and $A^{(j)}_t$ enters Eq.~\ref{eq:ppo}. The weights average to one over each
response, so a response keeps its total credit under GDPO and only its
placement changes. Positive and negative advantages use the same weights.
During training, $\beta$ is raised from $0$ over the first $N$ steps, and when
several rewards concentrate on the same tokens, the reallocated part of a token’s advantage is capped at the magnitude of its uniform advantage
(Appendix~\ref{app:weighting}). The bottom of Figure~\ref{fig:replay-sql} sums the
weighted advantages of the three rewards on each token.

\section{Experiments}
\label{sec:experiments}

\begin{table*}[t]
\centering
\scriptsize
\setlength{\tabcolsep}{1.5pt}
\renewcommand{\arraystretch}{1.1}
\begin{tabularx}{\textwidth}{@{}l*{10}{>{\centering\arraybackslash}X}@{}}
\toprule
 & \multicolumn{5}{c}{Non-Live} & \multicolumn{5}{c}{Live} \\
\cmidrule(lr){2-6}\cmidrule(l){7-11}
Method & Simple & Multiple & Parallel & Par.\ Mult. & Overall & Simple & Multiple & Parallel & Par.\ Mult. & Overall \\
\midrule
\multicolumn{11}{@{}l}{\emph{Qwen2.5-1.5B-Instruct}} \\
Base & 1.75 & 8.50 & 13.00 & 25.00 & 12.06 & 6.59 & 9.59 & 18.75 & 20.83 & 9.33 \\
GRPO & 70.07$_{\pm0.69}$ & 86.40$_{\pm0.42}$ & 81.90$_{\pm1.08}$ & 73.80$_{\pm2.20}$ & 78.04$_{\pm0.84}$ & \textbf{76.43}$_{\pm0.84}$ & 62.18$_{\pm0.75}$ & \textbf{55.00}$_{\pm2.80}$ & 60.83$_{\pm6.32}$ & 64.80$_{\pm0.55}$ \\
GDPO & 70.32$_{\pm2.54}$ & 88.60$_{\pm0.74}$ & 82.60$_{\pm2.63}$ & 75.70$_{\pm2.08}$ & 79.30$_{\pm1.69}$ & 71.78$_{\pm4.89}$ & 66.15$_{\pm1.93}$ & 53.75$_{\pm5.59}$ & 64.17$_{\pm4.75}$ & 67.05$_{\pm2.40}$ \\
\oursrow \ours & \textbf{70.90}$_{\pm1.10}$ & \textbf{89.00}$_{\pm0.50}$ & \textbf{83.10}$_{\pm1.56}$ & \textbf{77.20}$_{\pm2.25}$ & \textbf{80.05}$_{\pm1.00}$ & 74.42$_{\pm2.07}$ & \textbf{67.52}$_{\pm0.81}$ & \textbf{55.00}$_{\pm2.80}$ & \textbf{65.83}$_{\pm3.49}$ & \textbf{68.66}$_{\pm0.92}$ \\
\midrule
\multicolumn{11}{@{}l}{\emph{Qwen2.5-3B-Instruct}} \\
Base & 44.92 & 58.50 & 16.50 & 25.00 & 36.23 & 39.92 & 50.81 & 6.25 & 20.83 & 47.67 \\
GRPO & 73.92$_{\pm0.60}$ & 92.33$_{\pm0.76}$ & 81.67$_{\pm2.57}$ & 79.33$_{\pm2.02}$ & 81.81$_{\pm0.76}$ & 64.73$_{\pm0.39}$ & \textbf{70.69}$_{\pm0.11}$ & \textbf{72.92}$_{\pm3.61}$ & 55.56$_{\pm2.41}$ & 69.31$_{\pm0.11}$ \\
GDPO & 73.83$_{\pm0.63}$ & \textbf{92.67}$_{\pm0.29}$ & \textbf{82.33}$_{\pm1.89}$ & \textbf{80.83}$_{\pm2.36}$ & 82.42$_{\pm0.50}$ & 64.34$_{\pm2.05}$ & 70.21$_{\pm0.33}$ & \textbf{72.92}$_{\pm3.61}$ & 55.56$_{\pm8.67}$ & 68.86$_{\pm0.41}$ \\
\oursrow \ours & \textbf{75.36}$_{\pm0.84}$ & 92.33$_{\pm0.29}$ & 82.17$_{\pm2.02}$ & 80.33$_{\pm2.84}$ & \textbf{82.55}$_{\pm1.17}$ & \textbf{66.15}$_{\pm0.45}$ & 70.59$_{\pm0.79}$ & \textbf{72.92}$_{\pm3.61}$ & \textbf{59.72}$_{\pm4.81}$ & \textbf{69.58}$_{\pm0.66}$ \\
\bottomrule
\end{tabularx}
\caption{BFCL-v4 single-turn AST accuracy (\%); subscripts are standard deviations over seeds. \ours has the best overall accuracy on both splits with both models.}
\label{tab:bfcl}
\end{table*}

\begin{figure*}[t]
  \centering
  \includegraphics[width=\textwidth]{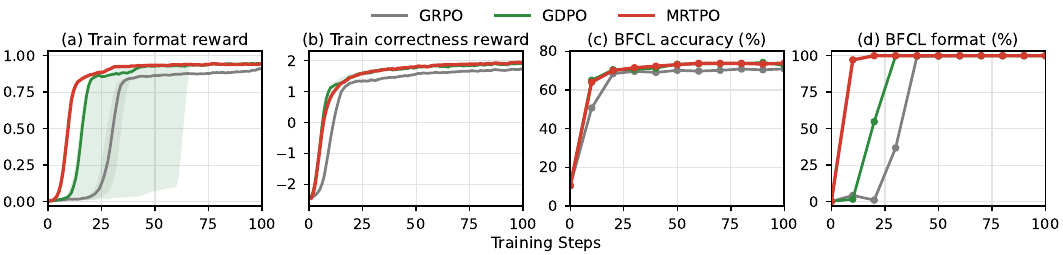}
  \caption{Tool calling on Qwen2.5-1.5B-Instruct over training: (a, b) training format and correctness rewards, median over seeds with interquartile range; (c, d) BFCL-v4 accuracy and format rate, mean over seeds. The binary format reward is harder to learn, while the continuous correctness reward rises early for all methods. The median number of steps to a format reward of 0.8 is 13 for \ours, 19 for GDPO and 34 for GRPO. }
  \label{fig:toolcall}
\end{figure*}

\subsection{Experimental Setup}
\label{sec:setup}

We evaluate \ours on three settings: text-to-SQL, instruction following and
tool calling. The complete training and evaluation parameters are given in
Appendix~\ref{app:hparams}, and the prompts and verifier blocks in
Appendix~\ref{app:contexts}.

\noindent \textbf{Text-to-SQL.}
Given a question and a database schema, the model reasons in
\texttt{<think>} and writes a SQL query in \texttt{<answer>}. We train on the
Spider training set \citep{yu2018spider} with three binary rewards following
SQL-R1 \citep{ma2025sqlr1}: format, executability and execution match against
the reference query. The verifier blocks of \ours state the required format,
the SQLite error raised by the response's own query, and the reference query.
We train Qwen2.5-Coder-3B-Instruct and 7B-Instruct \citep{hui2024qwen25coder}
for 100 steps and report greedy execution accuracy, the fraction of questions
whose query returns the same result as the reference, on Spider dev and test,
Spider-Syn \citep{gan2021syn}, Spider-Realistic \citep{deng2021realistic},
Spider-DK \citep{gan2021dk} and BIRD dev \citep{li2023bird}.

\noindent \textbf{Instruction following.}
We train on the first two subsets of HIR-16K \citep{hir2025}, whose
constraints are programmatically verifiable. Each prompt
carries five to thirteen constraints, and each constraint is a binary reward. The context pair of a constraint is the prompt with and
without it. We train Qwen2.5-7B-Instruct \citep{qwen2025qwen25} and
Qwen3-4B-Instruct-2507 \citep{yang2025qwen3} for 300 steps and report strict
prompt-level accuracy (all constraints of a prompt satisfied) and
instruction-level accuracy (fraction of constraints satisfied) on IFEval
\citep{zhou2023ifeval} and IFBench \citep{pyatkin2025ifbench}, and accuracy on
MulDimIF \citep{ye2025muldimif}, and, on AdvancedIF \citep{he2025advancedif}, whose rubrics are written by humans, the fraction of responses that satisfy all rubric items (Overall) and the fraction of rubric items satisfied (Rubric), with Qwen3.6-35B-A3B as the judge. All metrics are averaged over five samples per prompt.

\noindent \textbf{Tool calling.}
Following ToolRL \citep{qian2025toolrl}, the model selects tools and fills in
their arguments. We train on the RLLA data with a format reward and a
correctness reward that compares tool names, parameter names and values with
the reference calls. The verifier blocks of \ours state the required format
and list the reference calls. We train Qwen2.5-1.5B-Instruct and 3B-Instruct
\citep{qwen2025qwen25} for 100 steps and report the official AST accuracy on
the single-turn Non-Live and Live categories of BFCL-v4
\citep{patil2025bfcl}.

\noindent \textbf{Baselines.}
On all tasks we compare with GRPO \citep{shao2024deepseekmath} and GDPO
\citep{liu2026gdpo}. Most concurrent multi-reward methods reweight the rewards
when aggregating them; we select SAW \citep{he2026saw} as their representative.
Since reweighting is orthogonal to \ours, we compare with SAW on top of GDPO
and also apply it on top of \ours. On instruction following the rewards
differ across prompts, so such reweighting does not apply, and we instead
compare with CoRT \citep{zhang2026cort}, a token-level method. On text-to-SQL,
where reference reasoning chains are available \citep{li2025omnisql}, we also
compare with on-policy self-distillation (OPSD) \citep{zhao2026opsd}. \ours
uses $\beta=0.5$ on all tasks, and all results are averaged over independent
training runs.

\subsection{Main Results}
\label{sec:main}

\begin{figure*}[t]
  \centering
  \includegraphics[width=\textwidth]{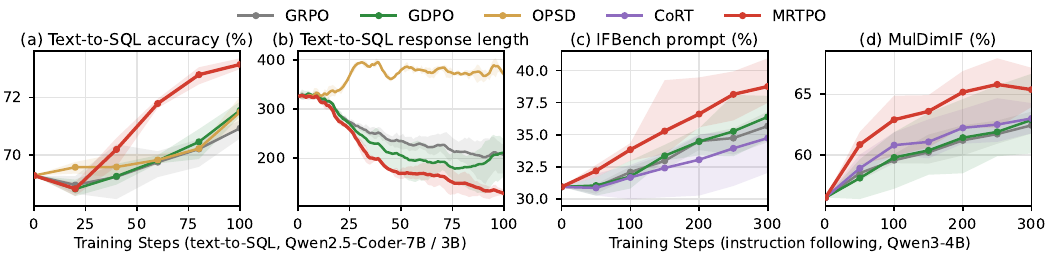}
  \caption{Training dynamics. (a) Text-to-SQL execution accuracy averaged over six benchmarks (Qwen2.5-Coder-7B-Instruct); (b) training response length (Qwen2.5-Coder-3B-Instruct); (c, d) IFBench prompt-level and MulDimIF accuracy (Qwen3-4B-Instruct-2507). Lines are seed means; bands span the seeds.}
  \label{fig:sql-eff}
\end{figure*}

\begin{figure*}[t]
  \centering
  \includegraphics[width=\textwidth]{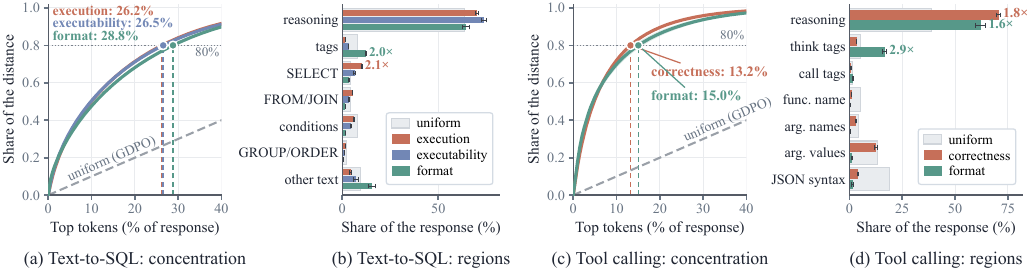}
  \caption{Where the replay distance falls. (a, b) Text-to-SQL with Qwen2.5-Coder-3B-Instruct; (c, d) tool calling with Qwen2.5-3B-Instruct. (a, c) Share of each reward's distance held by the top tokens of a response; (b, d) share on each part of the response. Gray marks the uniform assignment of GDPO.}
  \label{fig:credit}
\end{figure*}

\noindent \textbf{Does token-level assignment improve multi-reward RL?}
\ours has the best average performance on all three tasks and all six models
(Tables~\ref{tab:sql}, \ref{tab:hir} and~\ref{tab:bfcl}). \ours and GDPO
compute the same per-reward advantages and differ only in how each advantage
is assigned to the tokens, so the improvement comes from the token-level
assignment alone. \ours is orthogonal to response-level reweighting such as SAW and still outperforms it on text-to-SQL. Combined with SAW, it gives the best average with both models. OPSD gives its teacher more privileged information than \ours receives, a full reference solution
with the reasoning and the query, yet it stays below \ours on every
text-to-SQL benchmark. This shows that the policy needs only the information its verifiers already
use, not a full reference solution.

\noindent \textbf{When does it help most?}
The gain grows with the number of rewards in a response. Under a uniform
assignment every token receives the sum of all rewards' advantages, so most of
the credit on a token comes from rewards unrelated to it, and this share grows
with the number of rewards. Instruction following has five to thirteen
constraint rewards per prompt, and here \ours gains most, 2.4 and 3.2 points
on IFBench, whose constraints are not seen in training. Text-to-SQL has three
rewards, and \ours gains 1.3 and 1.6 points in average accuracy. Tool calling
has two, and \ours ends close to GDPO; its gain there is in speed.

\noindent \textbf{Does it learn faster?}
A reward decided by a few tokens receives only a small share of its advantage
on those tokens when the advantage is spread uniformly. The format reward of
tool calling, which its verifier checks on the tags alone, shows this most
clearly. After 10 steps, \ours produces the required format on 97\% of the
BFCL cases, compared with 2\% for GDPO and 4\% for GRPO
(Figure~\ref{fig:toolcall}d). The other tasks show the same pattern. With
Coder-7B, \ours passes the final text-to-SQL accuracy of GDPO at step 60 of
100, and with Qwen3-4B-Instruct-2507 it passes the final IFBench accuracy of
every baseline at step 200 of 300 (Figure~\ref{fig:sql-eff}a, c). In
text-to-SQL the responses also become 40\% shorter than those of GDPO with
Coder-3B (Figure~\ref{fig:sql-eff}b), because reasoning that does not affect
the query gets less execution advantage.

\begin{table}[t]
\centering
\small
\setlength{\tabcolsep}{3.5pt}
\resizebox{\columnwidth}{!}{%
\begin{tabular}{l rrr rrr}
\toprule
& \multicolumn{3}{c}{Qwen2.5-7B} & \multicolumn{3}{c}{Qwen3-4B} \\
\cmidrule(lr){2-4}\cmidrule(lr){5-7}
Constraint & Dist. & Tok. & Ratio & Dist. & Tok. & Ratio \\
\midrule
End phrase    & 39.6 & 4.4 & 9.1 & 41.4 & 4.0 & 10.4 \\
Title         & 24.5 & 3.2 & 7.6 & 25.0 & 3.3 & 7.5 \\
Quotation     & 12.3 & 1.1 & 11.1 & 19.6 & 1.1 & 17.4 \\
Start phrase  & 13.9 & 3.8 & 3.7 & -- & -- & -- \\
Keyword       & 8.6 & 2.7 & 3.2 & 10.2 & 3.1 & 3.3 \\
Postscript    & 34.3 & 14.2 & 2.4 & 41.8 & 15.5 & 2.7 \\
Bullet lines  & 89.4 & 86.3 & 1.0 & 68.8 & 45.7 & 1.5 \\
\midrule
All satisfied & 27.6 & 10.1 & 2.7 & 33.0 & 11.7 & 2.8 \\
All violated  & 17.7 & 9.5 & 1.9 & 7.4 & 1.4 & 5.1 \\
\bottomrule
\end{tabular}}
\caption{\textbf{Replay localizes each constraint.} For each satisfied constraint type, Dist.\ is the percentage of the constraint's replay distance on the part of the response its checker reads, Tok.\ is that part's percentage of the tokens, and Ratio is Dist.\ over Tok.; uniform assignment gives a ratio of 1. The last two rows pool all satisfied and all violated constraints. ``--'': fewer than five instances.}
\label{tab:hir-loc}
\end{table}

\noindent \textbf{Which tokens receive each reward's advantage?}
Without step-level annotations, the replay finds the tokens that decide each
reward from forward passes of the policy itself (Figure~\ref{fig:credit}). In
text-to-SQL, 80\% of a reward's distance falls on 26\% to 29\% of the tokens
(Figure~\ref{fig:credit}a). The execution reward puts 10\% of its distance on
the selected columns, twice their share of the tokens, and little on the
tags, while the format reward puts twice their share on the tags
(Figure~\ref{fig:credit}b). In tool calling, 80\% of a reward's distance falls
on 13\% to 15\% of the tokens (Figure~\ref{fig:credit}c). The format reward
gives the think tags three times their share of the tokens. The
correctness reward falls mostly on the reasoning that chooses the call, while
the function names and JSON syntax, which correct and incorrect calls share,
receive about a fifth of their share (Figure~\ref{fig:credit}d).

In instruction following, the checker of each constraint reads a known part
of the response; an end-phrase constraint, for example, asks the response to
finish with a given phrase, and its checker reads that phrase. We replay 192
HIR-16K responses from each of Qwen2.5-7B-Instruct and Qwen3-4B-Instruct-2507,
removing one constraint at a time (Table~\ref{tab:hir-loc}). A satisfied
constraint places about 30\% of its distance on the part its checker reads,
about a tenth of the response; an end phrase holds 40\% of the distance with
4\% of the tokens. A violated constraint moves its distance to where the
missing content should be, such as the opening token when quotation marks are
missing. The distances of different constraints on the same response are
weakly correlated, so each constraint receives its own assignment, and the
part of each advantage that follows relevance goes to the tokens where good
and bad responses differ.

\begin{table}[t]
\centering
\footnotesize
\setlength{\tabcolsep}{1.2pt}
\renewcommand{\arraystretch}{1.1}
\begin{tabularx}{\columnwidth}{@{}l*{7}{>{\centering\arraybackslash}X}@{}}
\toprule
Method & Dev & Test & Syn & Real. & DK & BIRD & Avg. \\
\midrule
GDPO & 79.21 & 80.62 & 72.21 & 74.87 & 67.29 & 55.11 & 71.55 \\
\oursrow \ours & \textbf{81.33} & \textbf{81.65} & \textbf{73.53} & \textbf{75.92} & \textbf{70.04} & \textbf{56.42} & \textbf{73.15} \\
\quad joint replay & 80.91 & 81.09 & 72.95 & \textbf{75.92} & 69.54 & 54.94 & 72.56 \\
\quad shuffled pos. & 77.92 & 80.41 & 71.21 & 75.26 & 66.54 & 54.85 & 71.03 \\
\bottomrule
\end{tabularx}
\caption{Ablations on text-to-SQL with Qwen2.5-Coder-7B-Instruct. Shuffling the weights within each response removes the gain over GDPO.}
\label{tab:ablation}
\end{table}

\begin{table}[t]
\centering
\footnotesize
\setlength{\tabcolsep}{0pt}
\renewcommand{\arraystretch}{1.1}
\begin{tabular*}{\columnwidth}{@{\extracolsep{\fill}}lcccccc@{}}
\toprule
& \multicolumn{2}{c}{IFEval} & \multicolumn{2}{c}{IFBench} & MulDimIF & \\
\cmidrule(lr){2-3}\cmidrule(lr){4-5}\cmidrule(lr){6-6}
$\beta$ & Prompt & Inst. & Prompt & Inst. & Acc. & Avg. \\
\midrule
0.25 & 81.59 & 87.31 & 35.83 & 37.91 & 53.69 & 59.26 \\
\oursrow 0.5 & 82.18 & 87.70 & \textbf{38.59} & \textbf{41.87} & 52.75 & \textbf{60.62} \\
0.75 & \textbf{82.49} & \textbf{88.04} & 36.08 & 38.67 & \textbf{55.04} & 60.06 \\
\bottomrule
\end{tabular*}
\caption{Allocation strength on instruction following with Qwen2.5-7B-Instruct. Assigning half of each advantage by relevance gives the highest average.}
\label{tab:beta-if}
\end{table}

\begin{table}[t]
\centering
\footnotesize
\setlength{\tabcolsep}{1.2pt}
\renewcommand{\arraystretch}{1.1}
\begin{tabularx}{\columnwidth}{@{}l*{7}{>{\centering\arraybackslash}X}@{}}
\toprule
$\beta$ & Dev & Test & Syn & Real. & DK & BIRD & Avg. \\
\midrule
0.25 & 81.24 & \textbf{83.05} & \textbf{73.69} & 76.18 & \textbf{71.54} & 55.87 & \textbf{73.59} \\
\oursrow 0.5 & \textbf{81.82} & 81.46 & 73.02 & \textbf{77.56} & 69.85 & 56.46 & 73.36 \\
0.75 & 80.85 & 81.97 & 72.73 & 74.41 & 70.04 & \textbf{56.92} & 72.82 \\
\bottomrule
\end{tabularx}
\caption{Allocation strength on text-to-SQL with Qwen2.5-Coder-7B-Instruct (one seed).}
\label{tab:beta-sql}
\end{table}

\noindent \textbf{Does the gain come from assigning each reward to its own tokens?}
Two ablations change only the assignment (Table~\ref{tab:ablation}).
Shuffling the token weights within each response keeps their values and
removes their positions, and the gain over GDPO disappears, so the gain depends
on where each advantage is assigned. Replaying all rewards together assigns
their summed advantage with one shared profile. It improves over GDPO but
stays below \ours, so the gain also depends on giving each reward its own
assignment. We set $\beta=0.5$ on all tasks instead of tuning it per task.
On two tasks we also tried 0.25 and 0.75 (Tables~\ref{tab:beta-if}
and~\ref{tab:beta-sql}). On instruction following, 0.5 gives the highest
IFBench accuracy and average, while 0.75 is higher on IFEval and MulDimIF; on
text-to-SQL, 0.25 gives a slightly higher average. The best $\beta$ thus
depends on the task, and tuning it per task could improve \ours further,
which we leave aside because of its compute cost.

\noindent \textbf{What does the replay cost?}
Replay adds forward passes of the policy. For each reward, \ours replays a
response only when that reward varies within its group, since otherwise its
advantage is zero. Such groups are common only early in training and quickly
become rare: in tool calling, the share of groups whose correctness varies
falls from 61\% in the first ten steps to about 10\%. Overall, 12\% to 21\% of
the response--reward pairs are replayed, and a training step takes 10\% to
20\% longer than with GDPO on the same GPUs.
\section{Conclusion}
\label{sec:conclusion}

We introduce \ours, which tells a multi-reward learner not only which rewards a response earned but also which tokens earned each of them. The key idea of \ours is to replay the policy’s own response under the prompt with and without one verifier’s information, such as a reference query, an execution error or a constraint. The tokens whose predictions change most are the ones this reward depends on, and \ours assigns more of the reward’s advantage to them, without step-level annotations. Across text-to-SQL, instruction following and tool calling, \ours consistently outperforms GRPO and GDPO on six models and reaches the final accuracy of GDPO in fewer steps. Any reward that comes with checkable information, such as a reference answer, an execution error or a rubric item, can be assigned to its tokens this way.
\section*{Limitations}
\ours requires that each reward's verification information can be written as
text and that adding or removing it changes the 
predictions from the policy.
Rewards without such information, for example response length, cannot be localized this way and 
shall 
keep 
uniform allocation %
in GDPO. 
Our experiments use models of
1.5B to 7B parameters with a single allocation strength ($\beta=0.5$) on all tasks. Larger models, rewards judged by an LLM such as rubric items, and adaptive choices of $\beta$ remain to be studied.

\section*{Ethics Statement}
This work studies credit assignment for reinforcement learning on publicly
available text-to-SQL, instruction-following and tool-calling datasets with programmatic verifiers. It does not involve human subjects or new data collection. More efficient optimization of verifiable rewards also makes it easier to optimize misspecified rewards.

\bibliography{references}

@article{shao2024deepseekmath,
  title={{DeepSeekMath}: Pushing the Limits of Mathematical Reasoning in Open Language Models},
  author={Shao, Zhihong and Wang, Peiyi and Zhu, Qihao and Xu, Runxin and Song, Junxiao and Bi, Xiao and Zhang, Haowei and Zhang, Mingchuan and Li, Y. K. and Wu, Y. and Guo, Daya},
  journal={arXiv preprint arXiv:2402.03300},
  year={2024}
}

@inproceedings{liu2026gdpo,
  title={{GDPO}: Group reward-Decoupled Normalization Policy Optimization for Multi-reward {RL} Optimization},
  author={Liu, Shih-Yang and Dong, Xin and Lu, Ximing and Diao, Shizhe and Belcak, Peter and Liu, Mingjie and Chen, Min-Hung and Yin, Hongxu and Wang, Yu-Chiang Frank and Cheng, Kwang-Ting and Choi, Yejin and Kautz, Jan and Molchanov, Pavlo},
  booktitle={International Conference on Machine Learning},
  year={2026}
}

@inproceedings{zhao2026opsd,
  title={Self-Distilled Reasoner: On-Policy Self-Distillation for Large Language Models},
  author={Zhao, Siyan and Xie, Zhihui and Liu, Mengchen and Huang, Jing and Pang, Guan and Chen, Feiyu and Grover, Aditya},
  booktitle={International Conference on Machine Learning},
  year={2026}
}

@inproceedings{hubotter2026sdpo,
  title={Reinforcement Learning via Self-Distillation},
  author={H{\"u}botter, Jonas and L{\"u}beck, Frederike and Behric, Lejs Deen and Baumann, Anton and Bagatella, Marco and Marta, Daniel and Hakimi, Ido and Shenfeld, Idan and Kleine Buening, Thomas and Guestrin, Carlos and Krause, Andreas},
  booktitle={International Conference on Machine Learning},
  year={2026}
}

@article{zhang2026cort,
  title={{CoRT}: Counterfactual Replay for Token-Level Rubric-Guided Policy Optimization},
  author={Zhang, Bo-Wen and He, Junwei and Wang, Wen and Lv, Song-Lin and Ma, Wentao and Lin, Rongyi and Zhong, Shuhan and Guo, Lan-Zhe},
  journal={arXiv preprint arXiv:2607.25659},
  year={2026}
}

@inproceedings{rtt2026,
  title={Rubrics to Tokens: Bridging Response-level Rubrics and Token-level Rewards in Instruction Following Tasks},
  author={Xu, Tianze and Zheng, Yanzhao and Lu, Pengrui and Ye, Lyumanshan and Wu, Yong and Zhang, Zhentao and Yu, Yuanqiang and Ma, Chao and Zhu, Jihuai and Liu, Pengfei and Dong, Baohua and Zhu, Hangcheng and Huang, Ruohui and Yu, Gang},
  booktitle={Conference on Language Modeling},
  year={2026}
}

@inproceedings{hir2025,
  title={Replay Failures as Successes: Sample-Efficient Reinforcement Learning for Instruction Following},
  author={Zhang, Kongcheng and Yao, Qi and Liu, Shunyu and Zhang, Wenjian and Cen, Min and Zhou, Yang and Fang, Wenkai and Zhao, Yiru and Lai, Baisheng and Song, Mingli},
  booktitle={International Conference on Machine Learning},
  year={2026}
}

@inproceedings{qian2025toolrl,
  title={{ToolRL}: Reward is All Tool Learning Needs},
  author={Qian, Cheng and Acikgoz, Emre Can and He, Qi and Wang, Hongru and Chen, Xiusi and Hakkani-T{\"u}r, Dilek and Tur, Gokhan and Ji, Heng},
  booktitle={Advances in Neural Information Processing Systems},
  year={2025}
}

@article{zhou2023ifeval,
  title={Instruction-Following Evaluation for Large Language Models},
  author={Zhou, Jeffrey and Lu, Tianjian and Mishra, Swaroop and Brahma, Siddhartha and Basu, Sujoy and Luan, Yi and Zhou, Denny and Hou, Le},
  journal={arXiv preprint arXiv:2311.07911},
  year={2023}
}

@inproceedings{pyatkin2025ifbench,
  title={Generalizing Verifiable Instruction Following},
  author={Pyatkin, Valentina and Malik, Saumya and Graf, Victoria and Ivison, Hamish and Huang, Shengyi and Dasigi, Pradeep and Lambert, Nathan and Hajishirzi, Hannaneh},
  booktitle={Advances in Neural Information Processing Systems Datasets and Benchmarks Track},
  year={2025}
}

@inproceedings{ye2025muldimif,
  title={{MulDimIF}: A Multi-Dimensional Constraint Framework for Evaluating and Improving Instruction Following in Large Language Models},
  author={Ye, Junjie and Huang, Caishuang and Chen, Zhuohan and Fu, Wenjie and Yang, Chenyuan and Yang, Leyi and Wu, Yilong and Wang, Peng and Zhou, Meng and Yang, Xiaolong and Gui, Tao and Zhang, Qi and Shi, Zhongchao and Fan, Jianping and others},
  booktitle={Proceedings of the 64th Annual Meeting of the Association for Computational Linguistics},
  year={2026}
}

@inproceedings{dai2024saferlhf,
  title={Safe {RLHF}: Safe Reinforcement Learning from Human Feedback},
  author={Dai, Josef and Pan, Xuehai and Sun, Ruiyang and Ji, Jiaming and Xu, Xinbo and Liu, Mickel and Wang, Yizhou and Yang, Yaodong},
  booktitle={International Conference on Learning Representations},
  year={2024}
}

@inproceedings{rame2023rewardedsoups,
  title={Rewarded Soups: Towards {P}areto-Optimal Alignment by Interpolating Weights Fine-Tuned on Diverse Rewards},
  author={Ram{\'e}, Alexandre and Couairon, Guillaume and Dancette, Corentin and Gaya, Jean-Baptiste and Shukor, Mustafa and Soulier, Laure and Cord, Matthieu},
  booktitle={Advances in Neural Information Processing Systems},
  year={2023}
}

@inproceedings{wu2023finegrained,
  title={Fine-Grained Human Feedback Gives Better Rewards for Language Model Training},
  author={Wu, Zeqiu and Hu, Yushi and Shi, Weijia and Dziri, Nouha and Suhr, Alane and Ammanabrolu, Prithviraj and Smith, Noah A. and Ostendorf, Mari and Hajishirzi, Hannaneh},
  booktitle={Advances in Neural Information Processing Systems},
  year={2023}
}

@inproceedings{lightman2023verify,
  title={Let's Verify Step by Step},
  author={Lightman, Hunter and Kosaraju, Vineet and Burda, Yura and Edwards, Harri and Baker, Bowen and Lee, Teddy and Leike, Jan and Schulman, John and Sutskever, Ilya and Cobbe, Karl},
  booktitle={International Conference on Learning Representations},
  year={2024}
}

@inproceedings{kazemnejad2024vineppo,
  title={{VinePPO}: Refining Credit Assignment in {RL} Training of {LLMs}},
  author={Kazemnejad, Amirhossein and Aghajohari, Milad and Portelance, Eva and Sordoni, Alessandro and Reddy, Siva and Courville, Aaron and Le Roux, Nicolas},
  booktitle={International Conference on Machine Learning},
  year={2025}
}

@inproceedings{agarwal2024gkd,
  title={On-Policy Distillation of Language Models: Learning from Self-Generated Mistakes},
  author={Agarwal, Rishabh and Vieillard, Nino and Zhou, Yongchao and Stanczyk, Piotr and Garea, Sabela Ramos and Geist, Matthieu and Bachem, Olivier},
  booktitle={International Conference on Learning Representations},
  year={2024}
}

@article{qwen2025qwen25,
  title={Qwen2.5 Technical Report},
  author={{Qwen Team}},
  journal={arXiv preprint arXiv:2412.15115},
  year={2024}
}

@inproceedings{ma2025sqlr1,
  title={{SQL-R1}: Training Natural Language to {SQL} Reasoning Model By Reinforcement Learning},
  author={Ma, Peixian and Zhuang, Xialie and Xu, Chengjin and Jiang, Xuhui and Chen, Ran and Guo, Jian},
  booktitle={Advances in Neural Information Processing Systems},
  year={2025}
}

@inproceedings{yu2018spider,
  title={Spider: A Large-Scale Human-Labeled Dataset for Complex and Cross-Domain Semantic Parsing and Text-to-{SQL} Task},
  author={Yu, Tao and Zhang, Rui and Yang, Kai and Yasunaga, Michihiro and Wang, Dongxu and Li, Zifan and Ma, James and Li, Irene and Yao, Qingning and Roman, Shanelle and Zhang, Zilin and Radev, Dragomir},
  booktitle={Proceedings of the 2018 Conference on Empirical Methods in Natural Language Processing},
  year={2018}
}

@inproceedings{gan2021syn,
  title={Towards Robustness of Text-to-{SQL} Models against Synonym Substitution},
  author={Gan, Yujian and Chen, Xinyun and Huang, Qiuping and Purver, Matthew and Woodward, John R. and Xie, Jinxia and Huang, Pengsheng},
  booktitle={Proceedings of the 59th Annual Meeting of the Association for Computational Linguistics},
  year={2021}
}

@inproceedings{deng2021realistic,
  title={Structure-Grounded Pretraining for Text-to-{SQL}},
  author={Deng, Xiang and Awadallah, Ahmed Hassan and Meek, Christopher and Polozov, Oleksandr and Sun, Huan and Richardson, Matthew},
  booktitle={Proceedings of the 2021 Conference of the North American Chapter of the Association for Computational Linguistics},
  year={2021}
}

@inproceedings{gan2021dk,
  title={Exploring Underexplored Limitations of Cross-Domain Text-to-{SQL} Generalization},
  author={Gan, Yujian and Chen, Xinyun and Purver, Matthew},
  booktitle={Proceedings of the 2021 Conference on Empirical Methods in Natural Language Processing},
  year={2021}
}

@inproceedings{li2023bird,
  title={Can {LLM} Already Serve as A Database Interface? {A} {BIg} Bench for Large-Scale Database Grounded Text-to-{SQL}s},
  author={Li, Jinyang and Hui, Binyuan and Qu, Ge and Yang, Jiaxi and Li, Binhua and Li, Bowen and Wang, Bailin and Qin, Bowen and Cao, Rongyu and Geng, Ruiying and Huo, Nan and Zhou, Xuanhe and Ma, Chenhao and Li, Guoliang and others},
  booktitle={Advances in Neural Information Processing Systems Datasets and Benchmarks Track},
  year={2023}
}

@article{hui2024qwen25coder,
  title={{Qwen2.5-Coder} Technical Report},
  author={Hui, Binyuan and Yang, Jian and Cui, Zeyu and Yang, Jiaxi and Liu, Dayiheng and Zhang, Lei and Liu, Tianyu and Zhang, Jiajun and Yu, Bowen and Lu, Keming and others},
  journal={arXiv preprint arXiv:2409.12186},
  year={2024}
}

@inproceedings{patil2025bfcl,
  title={The {B}erkeley Function Calling Leaderboard ({BFCL}): From Tool Use to Agentic Evaluation of Large Language Models},
  author={Patil, Shishir G. and Mao, Huanzhi and Yan, Fanjia and Ji, Charlie Cheng-Jie and Suresh, Vishnu and Stoica, Ion and Gonzalez, Joseph E.},
  booktitle={International Conference on Machine Learning},
  year={2025}
}

@article{guo2025deepseekr1,
  title={{DeepSeek-R1} Incentivizes Reasoning in {LLMs} Through Reinforcement Learning},
  author={Guo, Daya and Yang, Dejian and Zhang, Haowei and Song, Junxiao and Wang, Peiyi and others},
  journal={Nature},
  volume={645},
  pages={633--638},
  year={2025}
}

@article{jiang2026dvao,
  title={{DVAO}: Dynamic Variance-Adaptive Advantage Optimization for Multi-Reward Reinforcement Learning},
  author={Jiang, Guochao and Song, Jingyi and Quan, Guofeng and Hao, Chuzhan and Liu, Guohua and Zhang, Yuewei},
  journal={arXiv preprint arXiv:2605.25604},
  year={2026}
}

@article{he2026saw,
  title={{SAW}: Stage-Aware Dynamic Weighting for Multi-Objective Reinforcement Learning in Large Language Models},
  author={He, Yuchen and Bi, Baolong and Liu, Shenghua and Liao, Huaming and Ge, Yuyao and Wan, Bolin and Tong, Siqian and Chen, Juan and Guo, Jiafeng and Cheng, Xueqi},
  journal={arXiv preprint arXiv:2606.07705},
  year={2026}
}

@article{wang2026samrpo,
  title={Learn What's Left, Not What's Mastered: Saturation Aware Advantage Reweighting for Multi-Reward Policy Optimization},
  author={Wang, Yixuan and Chen, Yifei and Zhang, Haichao and Luo, Haozheng and Wu, Xander and Ni, Jie and Fu, Yun and Vasconcelos, Nuno and Li, Yijiang},
  journal={arXiv preprint arXiv:2608.16072},
  year={2026}
}

@article{liu2026gd2po,
  title={{GD$^2$PO}: Mitigating Multi-Reward Conflicts via Group-Dynamic reward-Decoupled Policy Optimization},
  author={Liu, Haotian and Liu, Yihao and Ni, Jingwei and Huang, Siyuan and Liu, Xinpeng and Cheng, Pengyu and Song, Jiajun and Ding, Ruijin and Li, Junfeng and Yu, Zhechao and Zhou, Mengyu and Xu, Hongteng and Jiang, Xiaoxi and Jiang, Guanjun},
  journal={arXiv preprint arXiv:2606.16771},
  year={2026}
}

@article{montero2026rvpo,
  title={{RVPO}: Risk-Sensitive Alignment via Variance Regularization},
  author={Montero, Ivan and Jurczyk, Tomasz and Dhingra, Bhuwan},
  journal={arXiv preprint arXiv:2605.05750},
  year={2026}
}

@inproceedings{wang2025highentropy,
  title={Beyond the 80/20 Rule: High-Entropy Minority Tokens Drive Effective Reinforcement Learning for {LLM} Reasoning},
  author={Wang, Shenzhi and Yu, Le and Gao, Chang and Zheng, Chujie and Liu, Shixuan and Lu, Rui and Dang, Kai and Chen, Xionghui and Yang, Jianxin and Zhang, Zhenru and Liu, Yuqiong and Yang, An and Zhao, Andrew and Yue, Yang and Song, Shiji and Yu, Bowen and Huang, Gao and Lin, Junyang},
  booktitle={Advances in Neural Information Processing Systems},
  year={2025}
}

@inproceedings{cheng2025entropy,
  title={Reasoning with Exploration: An Entropy Perspective},
  author={Cheng, Daixuan and Huang, Shaohan and Zhu, Xuekai and Dai, Bo and Zhao, Wayne Xin and Zhang, Zhenliang and Wei, Furu},
  booktitle={Proceedings of the AAAI Conference on Artificial Intelligence},
  year={2026}
}

@inproceedings{wang2025empg,
  title={Harnessing Uncertainty: Entropy-Modulated Policy Gradients for Long-Horizon {LLM} Agents},
  author={Wang, Jiawei and Liu, Jiacai and Fu, Yuqian and Li, Yingru and Wang, Xintao and Lin, Yuan and Yue, Yu and Zhang, Lin and Wang, Yang and Wang, Ke},
  booktitle={International Conference on Machine Learning},
  year={2026}
}

@article{yang2025qwen3,
  title={Qwen3 Technical Report},
  author={Yang, An and Li, Anfeng and Yang, Baosong and Zhang, Beichen and Hui, Binyuan and Zheng, Bo and Yu, Bowen and Gao, Chang and Huang, Chengen and Lv, Chenxu and others},
  journal={arXiv preprint arXiv:2505.09388},
  year={2025}
}

@article{he2025advancedif,
  title={{AdvancedIF}: Rubric-Based Benchmarking and Reinforcement Learning for Advancing {LLM} Instruction Following},
  author={He, Yun and Li, Wenzhe and Zhang, Hejia and Li, Songlin and Mandyam, Karishma and Khosla, Sopan and Xiong, Yuanhao and Wang, Nanshu and Peng, Xiaoliang and Li, Beibin and others},
  journal={arXiv preprint arXiv:2511.10507},
  year={2025}
}

@article{li2025omnisql,
  title={{OmniSQL}: Synthesizing High-Quality Text-to-{SQL} Data at Scale},
  author={Li, Haoyang and Wu, Shang and Zhang, Xiaokang and Huang, Xinmei and Zhang, Jing and Jiang, Fuxin and Wang, Shuai and Zhang, Tieying and Chen, Jianjun and Shi, Rui and Chen, Hong and Li, Cuiping},
  journal={Proceedings of the VLDB Endowment},
  volume={18},
  number={11},
  pages={4695--4709},
  year={2025}
}

@article{zheng2026makedara,
  title={Make Sparse Rewards Count: Density-Aware Reward Aggregation for Multi-Reward {RL}},
  author={Zheng, Tong and Zhai, Skylar and Cheng, Zhan and Sha, TianMing and Huang, Youling and Zhou, Shuo and Qi, Shaotong and Liang, Jingcheng and Ding, Xuwei and Xu, Pengcheng},
  journal={arXiv preprint arXiv:2610.00574},
  year={2026}
}

\appendix
\section{Relation to Reweighting Methods}
\label{app:reweight}

Algorithm~\ref{alg:mrtpo} compares one training step of GDPO and \ours. Most
concurrent multi-reward methods change only how the per-reward advantages are
combined into one advantage per response. They weight each reward by batch or
group statistics, mask or weight responses, or replace the sum with a soft
minimum, and every token of a response receives the same advantage
(Table~\ref{tab:reweight}). \ours instead changes how each advantage is
assigned to the tokens, so it is orthogonal to all of these methods. In this
paper we combine it with SAW as an example (Table~\ref{tab:sql}).

\definecolor{newstep}{RGB}{31,86,170}
\newcommand{\added}[1]{\textcolor{newstep}{#1}}
\begin{algorithm*}[t]
\small
\begin{minipage}[t]{0.47\textwidth}
\textbf{GDPO}\par\vspace{2pt}\hrule\vspace{3pt}
\begin{algorithmic}[1]
\For{each prompt $P_0$ in the batch}
  \State Sample $G$ responses $y^{(1)},\dots,y^{(G)}$ from $\pi_{\theta_{\mathrm{old}}}$
  \State Score every response with each verifier: $r^{(j)}_k$
  \State Standardize each reward within the group:
    \Statex \hspace{\algorithmicindent}$A^{(j)}_k=\mathcal{G}\big(r^{(j)}_k\big)$
  \State Sum the rewards: $S^{(j)}=\sum_k A^{(j)}_k$
  \State Give every token the same advantage:
    \Statex \hspace{\algorithmicindent}$S^{(j)}_t=S^{(j)}$ for all $t$
\EndFor
\State Rescale: $A^{(j)}_t=S^{(j)}_t/\mathrm{std}(S)$
\State Update $\pi_\theta$ with the clipped objective
\end{algorithmic}
\end{minipage}\hfill
\begin{minipage}[t]{0.5\textwidth}
\textbf{\ours}\par\vspace{2pt}\hrule\vspace{3pt}
\begin{algorithmic}[1]
\For{each prompt $P_0$ in the batch}
  \State Sample $G$ responses $y^{(1)},\dots,y^{(G)}$ from $\pi_{\theta_{\mathrm{old}}}$
  \State Score every response with each verifier: $r^{(j)}_k$
  \State Standardize each reward within the group:
    \Statex \hspace{\algorithmicindent}$A^{(j)}_k=\mathcal{G}\big(r^{(j)}_k\big)$
  \State \added{For each reward $k$, find its relevant tokens:}
    \Statex \hspace{\algorithmicindent}\added{replay $y^{(j)}$ after $P_{\mathrm{all}}$ and after $P_{\setminus k}$}
    \Statex \hspace{\algorithmicindent}\added{$q_{k,t}=$ JS distance of the two predictions}
    \Statex \hspace{\algorithmicindent}\added{$w_{k,t}=(1-\beta)+\beta\,q_{k,t}/\bar q_k$}
  \State Sum the rewards \added{on each token}:
    \Statex \hspace{\algorithmicindent}$S^{(j)}_t=\sum_k A^{(j)}_k\,\added{w_{k,t}}$
\EndFor
\State Rescale: $A^{(j)}_t=S^{(j)}_t/\mathrm{std}(S)$
\State Update $\pi_\theta$ with the clipped objective
\end{algorithmic}
\end{minipage}
\caption{One training step of GDPO (left) and \ours (right); the steps \ours adds are in blue
(Eqs.~\ref{eq:contexts}--\ref{eq:token-adv}). Replay reuses $\pi_{\theta_{\mathrm{old}}}$ and samples nothing new;
the warm-up of $\beta$ and the reallocation bound are in Appendix~\ref{app:weighting}.}
\label{alg:mrtpo}
\end{algorithm*}

\begin{table*}[t]
\centering
\small
\setlength{\tabcolsep}{6pt}
\renewcommand{\arraystretch}{1.15}
\begin{tabularx}{\textwidth}{@{}l>{\raggedright\arraybackslash}Xl@{}}
\toprule
Method & What changes in the GDPO step & Token advantage $S^{(j)}_t$ \\
\midrule
GDPO \citep{liu2026gdpo} & nothing; the per-reward advantages are summed & $\sum_k A^{(j)}_k$ for all $t$ \\
SAW \citep{he2026saw} & each reward is weighted by its coefficient of variation in the batch & $\sum_k \lambda_k A^{(j)}_k$ for all $t$ \\
DARA \citep{zheng2026makedara} & rewards that vary in fewer groups get a larger, capped weight on their positive advantages & $\sum_k \big(A^{(j)}_k+(\lambda_k-1)[A^{(j)}_k]_+\big)$ for all $t$ \\
DVAO \citep{jiang2026dvao} & each reward is weighted by its standard deviation within the group, normalized across rewards & $\sum_k \lambda_k A^{(j)}_k$ for all $t$ \\
SA-MRPO \citep{wang2026samrpo} & each reward is discounted by how saturated it is in the batch & $\sum_k \lambda_k A^{(j)}_k$ for all $t$ \\
GD$^2$PO \citep{liu2026gd2po} & responses whose per-reward advantages disagree in sign are dropped, and each prompt is scaled by its share of kept responses & $\kappa\,\delta^{(j)}\sum_k A^{(j)}_k$ for all $t$ \\
RVPO \citep{montero2026rvpo} & the rewards are combined by a soft minimum, which penalizes their variance & $\mathrm{softmin}_k\,A^{(j)}_k$ for all $t$ \\
\midrule
\oursrow \ours & each reward's advantage is assigned to tokens by relevance & $\sum_k A^{(j)}_k\,w_{k,t}$ \\
\bottomrule
\end{tabularx}
\caption{Where each method changes the GDPO step. The other methods act on the response, so all tokens of a response receive the same advantage; \ours sets a weight per reward and token and can be combined with them.}
\label{tab:reweight}
\end{table*}

\section{Training and Evaluation Parameters}
\label{app:hparams}

Table~\ref{tab:hp} lists the training and evaluation
parameters of the three tasks. Reported results are averaged over independent training runs. All tasks use AdamW ($\beta_1=0.9$, $\beta_2=0.999$), and \ours
always uses $\beta=0.5$ with an $N=10$-step smoothstep warm-up and the
reallocation bound of Eq.~\ref{eq:bound}. GRPO
standardizes the aggregated reward within each group, without batch-level
rescaling.

\begin{table*}[t]
\centering
\footnotesize
\setlength{\tabcolsep}{4pt}
\renewcommand{\arraystretch}{1.1}
\begin{tabularx}{\textwidth}{@{}l>{\raggedright\arraybackslash}X>{\raggedright\arraybackslash}X>{\raggedright\arraybackslash}X@{}}
\toprule
Parameter & Text-to-SQL & Instruction following & Tool calling \\
\midrule
Models & Qwen2.5-Coder-\{3B, 7B\}-Instruct & Qwen2.5-7B-Instruct; Qwen3-4B-Instruct-2507 & Qwen2.5-\{1.5B, 3B\}-Instruct \\
Train prompts & Spider train, 6,915 & HIR-16K, first two subsets, 6,998 & RLLA, 3,920 (80 for validation) \\
Rewards & format, executable, execution; $\{0,1\}$ & one per constraint, $\{0,1\}$; 5--13 & format $\{0,1\}$; correctness $[-3,3]$ \\
Training steps & 100 & 300 & 100 \\
Prompts per step / $G$ & 32 / 8 & 64 / 8 & 512 / 4 \\
Rollout temp.\ / top-$p$ / top-$k$ & 1.1 / 1.0 / none & 1.0 / 0.99 / 100 & 1.0 / 1.0 / none \\
Max prompt / response length & 4,096 / 2,048 & 2,048 / 4,096 & 2,048 / 1,024 \\
Learning rate & $10^{-6}$, 15-step linear warm-up & $10^{-6}$, 15-step linear warm-up & $10^{-6}$, constant \\
Weight decay / grad.\ clip & 0.1 / 1.0 & 0.1 / 1.0 & 0.01 / 1.0 \\
Updates per step & 1 & 1 & 4 (mini-batch of 128 prompts) \\
PPO clip $[\varepsilon_l,\varepsilon_h]$ & [0.2, 0.2] & [0.2, 0.27] & [0.2, 0.2] \\
Entropy coefficient & 0 & 0 & 0.001 \\
KL penalty & in loss, low-var KL, 0.001 & none & in reward, coefficient 0.001 \\
Batch rescaling (GDPO, \ours) & per-response std & per-response std & token-level whitening \\
Validation & -- & -- & 80 prompts, greedy, every 10 steps \\
Number of seeds & 3 & 5 & 5 \\
Hardware & 8 H20 GPUs & 8 H200 GPUs & 4 A100 GPUs \\
\midrule
Evaluation & -- & -- & BFCL-v4 single-turn (gorilla \texttt{6ea57973}), step 100 \\
Evaluation decoding & greedy & temp.\ 0.7, top-$p$ 0.8, top-$k$ 20 & temp.\ 0.6, top-$p$ 0.95, seed 0 \\
Samples / max length & -- & 5 / 4,096 & -- \\
Max new tokens / context & -- & -- & 8,192 / 32,768 \\
\midrule
OPSD teacher & initial model with reference solution & -- & -- \\
OPSD divergence & forward KL on teacher top-64 tokens & -- & -- \\
OPSD per-token clip & 2.0 & -- & -- \\
\bottomrule
\end{tabularx}
\caption{Training and evaluation parameters of the three tasks.}
\label{tab:hp}
\end{table*}

\paragraph{Text-to-SQL.} We drop 3 Spider training questions whose reference
query fails to execute and 82 whose prompt exceeds 4,096 tokens. The reference
solutions of OPSD are the chain-of-thought solutions that OmniSQL releases for
the Spider training set. OPSD shares the rollout and optimizer settings of the
other methods, and its teacher is the initial model whose context shows the
reference solution as an earlier answer to the same question, followed by the
original prompt. For SAW, each reward's group advantage is multiplied
by its coefficient of variation in the batch, normalized so that the three
weights sum to three.

\paragraph{Instruction following.} We use the first two subsets of HIR-16K,
6,998 prompts with 43,496 constraints, because all their constraints are
checked by programs, while the other subsets need an LLM judge.

\paragraph{Tool calling.} We follow the public GDPO implementation. Reference
argument values written as numeric or boolean strings are converted to the
types declared in the tool schema.

\section{Replay contexts}
\label{app:contexts}

\paragraph{Text-to-SQL.} The format block is
\begin{promptbox}{Text-to-SQL: format block}\small\ttfamily
For the final user request, respond in this form:\\
<think>...</think>\\
<answer>\\
\textasciigrave\textasciigrave\textasciigrave{}sql\\
...\\
\textasciigrave\textasciigrave\textasciigrave{}\\
</answer>\\
Use exactly one <think> block followed by exactly one <answer> block, and put
the final SQL query in a \textasciigrave\textasciigrave\textasciigrave{}sql code block inside <answer>.
\end{promptbox}
The execution block states the reference query, ``The reference query for this
question is: \texttt{<query>}.'' The executability block of a failing response
reports its own error, ``When this query was executed on this database, SQLite
raised: \texttt{<error>}.'' A response without a SQL block in its answer
receives ``This response contains no SQL code block inside <answer>.''

\paragraph{Instruction following.} For a prompt with constraints
$c_1,\dots,c_M$, $P_{\setminus k}$ is produced by the same HiR concatenation template with
$c_k$ omitted. For
example, removing the end-phrase constraint from
\begin{promptbox}{Instruction following: removing the end-phrase constraint}\small
Provide a detailed response to the following query: ``Explain the importance of
maintaining a balanced diet for athletes.'' Your ENTIRE response should be in en
language \dots{} Your answer must contain exactly 3 bullet points \dots{} At the
end of your response, please explicitly add a postscript starting with P.S.
\textbf{Finish your response with this exact phrase Is there anything else I can
help with? No other words should follow this phrase.} Wrap your entire response
with double quotation marks.
\end{promptbox}
deletes only the bold sentence.

\paragraph{Tool calling.} Both blocks are appended to the last user message,
separated by a blank line, correctness first. The format block for a tool-call
example is
\begin{promptbox}{Tool calling: format block}\small\ttfamily
For the final user request, respond in this form:\\
<think>...</think>\\
<tool\_call>\\
...\\
</tool\_call>\\
Use one <tool\_call> block, place its content between the opening and closing
tags on separate lines, and end the answer with </tool\_call>. Do not add a
<response> block.
\end{promptbox}
and the analogous block with \texttt{<response>} for direct-response examples.
The correctness block lists the reference calls:
\begin{promptbox}{Tool calling: correctness block}\small\ttfamily
The correctness reference for the final user request contains these ordered
operation:\\
1. Operation name: `reverse\_string`.\\
\hspace*{1em}Parameter `text`: "Python programming".\\
The operation order, names, parameter names, and parameter values above are the
complete correctness reference.
\end{promptbox}

\begin{figure*}[t]
  \centering
  \includegraphics[width=\textwidth]{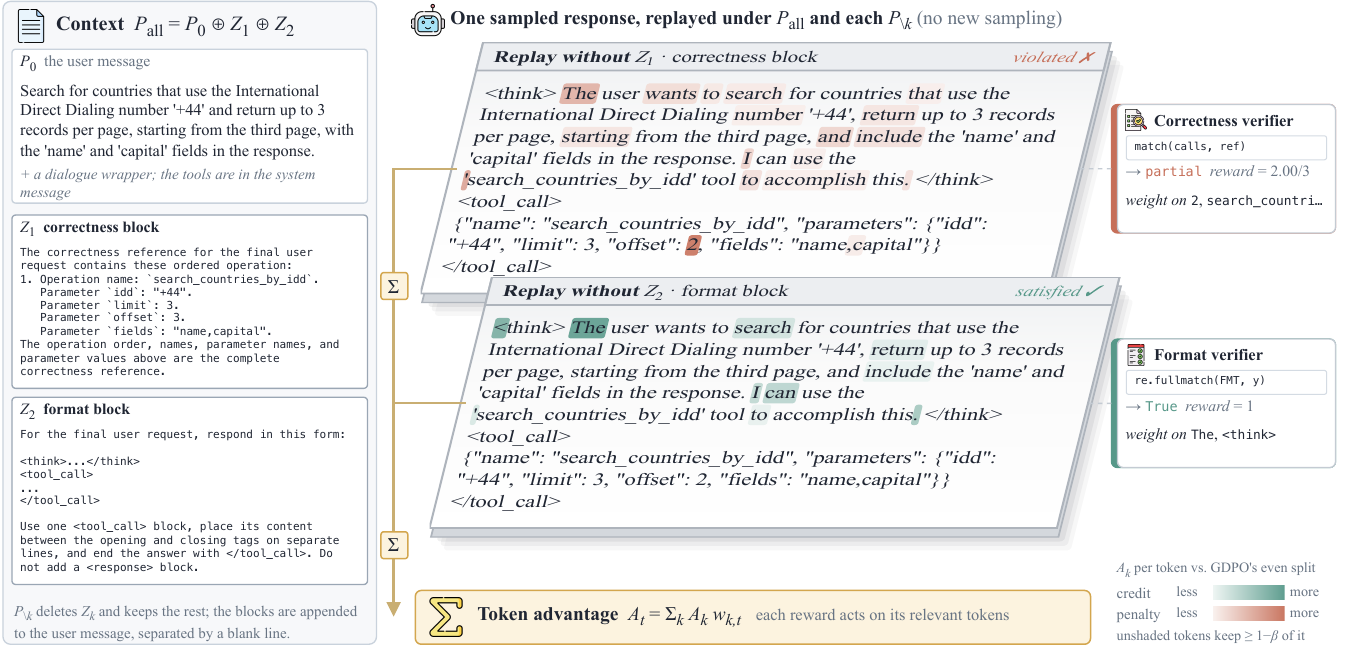}
  \caption{\textbf{Leave-one-out replay on a tool-calling response.} The response
  sets \texttt{offset} to 2; the reference has 3. The layout follows
  Figure~\ref{fig:replay-sql}.}
  \label{fig:replay-tool}
\end{figure*}

\begin{figure*}[t]
  \centering
  \includegraphics[width=\textwidth]{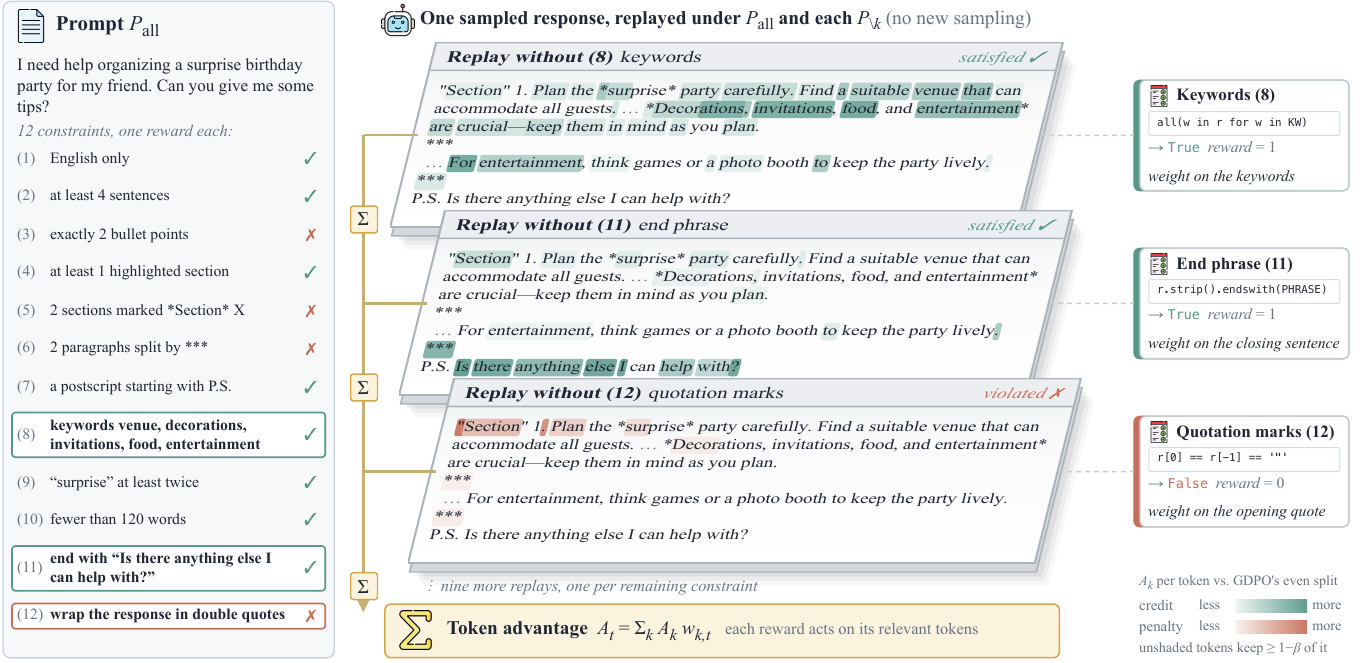}
  \caption{\textbf{Leave-one-out replay on an instruction-following response}
  with 12 constraints, three of whose replays are shown. Layout and shading
  follow Figure~\ref{fig:replay-sql}.}
  \label{fig:hir-example}
\end{figure*}

\paragraph{Replay examples.}
Figures~\ref{fig:replay-sql}, \ref{fig:replay-tool} and~\ref{fig:hir-example}
show the replays of one training response for text-to-SQL, tool calling and
instruction following.

\section{Relevance and Weighting Details}
\label{app:weighting}

\paragraph{Jensen--Shannon divergence.}
With $m_{k,t}=\frac12(p^{\mathrm{all}}_t+p^{\setminus k}_t)$, the divergence in
Eq.~\ref{eq:q} is computed over the full vocabulary as
\begin{equation*}
\begin{aligned}
\JS\big(p^{\mathrm{all}}_t\,\|\,p^{\setminus k}_t\big)
&=\tfrac12 D_{\mathrm{KL}}\big(p^{\mathrm{all}}_t\,\|\,m_{k,t}\big)\\
&\quad+\tfrac12 D_{\mathrm{KL}}\big(p^{\setminus k}_t\,\|\,m_{k,t}\big).
\end{aligned}
\end{equation*}
It is symmetric and at most $\ln 2$, so $q_{k,t}\in[0,1]$. Comparing full
distributions rather than the probability of the sampled token alone also
captures mass that moves among competing candidates while the sampled token
keeps its probability.

\paragraph{Warm-up.}
At training step $n$, Eq.~\ref{eq:weight} uses $\beta_n=\beta\,\lambda_n$ in
place of $\beta$, with the smoothstep $\lambda_n=3u^2-2u^3$ and
$u=\min(n/N,1)$, which rises from $0$ to $1$ over the first $N$ steps with zero
slope at both ends.

\paragraph{Bounding the reallocation.}
For response $j$, write the token advantage of Eq.~\ref{eq:token-adv} before batch
rescaling as $S^{(j)}_t=S^{(j)}+R^{(j)}_t$, where $S^{(j)}=\sum_k A^{(j)}_k$ is
the uniform GDPO advantage and $R^{(j)}_t=\sum_k A^{(j)}_k\,(w^{(j)}_{k,t}-1)$ is
the reallocated part, which has zero mean over the response. Over the valid
tokens of a training batch we compute
\begin{equation}
s=\min\Big(1,\sqrt{\textstyle\sum_j T_j\,(S^{(j)})^2\big/\sum_j\sum_t (R^{(j)}_t)^2}\,\Big)
\label{eq:bound}
\end{equation}
and use $S^{(j)}+s\,R^{(j)}_t$ in place of $S^{(j)}_t$. This is the same as
replacing every weight $w_{k,t}$ by $1+s\,(w_{k,t}-1)$, which keeps each
response's mean advantage and bounds the second moment of the token advantage
by twice that of the uniform advantage.

\end{document}